\documentclass[conference]{IEEEtran}
\IEEEoverridecommandlockouts
\usepackage{tikz}
\usetikzlibrary{arrows.meta,positioning,fit,shapes.misc}

\usepackage{cite}
\usepackage{amsmath,amssymb,amsfonts}
\usepackage{algorithmic}
\usepackage{graphicx}
\usepackage{booktabs} 
\usepackage{textcomp}
\usepackage{xcolor}
\usepackage{booktabs}
\usepackage{booktabs,makecell,placeins}
\usepackage{booktabs,makecell,placeins,stfloats}
\def\BibTeX{{\rm B\kern-.05em{\sc i\kern-.025em b}\kern-.08em
    T\kern-.1667em\lower.7ex\hbox{E}\kern-.125emX}}
\begin{document}

\title{Renaissance of RNNs in Streaming Clinical Time Series: Compact Recurrence Remains Competitive with Transformers\\

}

\author{
\IEEEauthorblockN{Ran Tong}
\IEEEauthorblockA{\textit{University of Texas at Dallas}\\
United States\\
rxt200012@utdallas.edu}
\and
\IEEEauthorblockN{Jiaqi Liu }
\IEEEauthorblockA{\textit{Independent Researcher}\\
United States\\
jackyliu9747@gmail.com}
\and
\IEEEauthorblockN{Su Liu}
\IEEEauthorblockA{\textit{Georgia Institute of Technology}\\
United States\\
sliu792@gatech.edu}
\and
\IEEEauthorblockN{Xin Hu }
\IEEEauthorblockA{\textit{University of Michigan- Ann Arbor}\\
United States\\
hsinhu@umich.edu}
\and
\IEEEauthorblockN{Lanruo Wang}
\IEEEauthorblockA{\textit{University of Texas at Dallas}\\
United States\\
lxw220021@utdallas.edu}

}
\maketitle

\begin{abstract}
We present a compact, strictly causal benchmark for streaming clinical time series on the MIT--BIH Arrhythmia Database using per-second heart rate. Two tasks are studied under record-level, non-overlapping splits: near-term tachycardia risk (next ten seconds) and one-step heart rate forecasting. We compare a  GRU-D (RNN) and a Transformer under matched training budgets against strong non-learned baselines. Evaluation is calibration-aware for classification and proper for forecasting, with temperature scaling and grouped bootstrap confidence intervals. On MIT-BIH, GRU-D slightly surpasses the Transformer for tachycardia risk, while the Transformer clearly lowers forecasting error relative to GRU-D and persistence. Our results show that, in longitudinal monitoring, model choice is task-dependent: compact RNNs remain competitive for short-horizon risk scoring, whereas compact Transformers deliver clearer gains for point forecasting.
\end{abstract}

\begin{IEEEkeywords}
Clinical Time Series, Longitudinal Data, Heart Rate Forecasting, Tachycardia Classification, RNN (GRU-D), Transformer, PhysioNet MIT-BIH Arrhythmia
\end{IEEEkeywords}

\section{Introduction and Related Work}

Recurrent neural networks (RNNs) model temporal dependencies by evolving a latent state as observations unfold \cite{b1}. Gated variants—long short-term memory (LSTM) and gated recurrent units (GRUs)—address vanishing gradients and are widely used in speech, language, and biosignals \cite{b2,b12}. For clinical time series, RNNs are appealing because they operate causally, support streaming inference with tight latency budgets, and admit principled handling of missingness; GRU-D learns feature- and state-wise decays toward clinically plausible defaults and is a standard baseline for physiologic data \cite{b3}. Extensions such as time-aware LSTMs and attention over recurrent states target irregular sampling and clinician-facing explanations \cite{b17,b16}. Continuous-time formulations (latent ODEs, GRU-ODE-Bayes) explicitly model event times \cite{b18,b25}, while temporal convolutional networks can be competitive when fixed receptive fields suffice \cite{b22}. In adjacent forecasting domains, multi-frequency data-fusion architectures have improved predictive accuracy \cite{xiao2025carbon}. Classical encoder–decoder training with additive attention established content-based weighting in sequence transduction before modern Transformers \cite{b15,b13}. Beyond purely sequence models, longitudinal clinical trajectories are also modeled with statistical and neural mixed-effects approaches \cite{tong2025predicting}, which contextualizes our benchmark’s focus on compact causal encoders for streaming monitoring.

Transformers reframe sequence learning around content-based self-attention \cite{b4} and now dominate language \cite{b5,b6} and vision \cite{b7}, even as lightweight multimodal systems show that compact architectures can be effective under tight compute and data budgets \cite{liu2025memeblip2}. Efficiency-oriented attention reduces quadratic cost and enables longer contexts \cite{b24}; time-series–focused designs tailor inductive biases for forecasting (e.g., sparse attention and series decomposition) \cite{b19,b20}. Clinical applications leverage large pretrained language models for biomedical text and notes, offering strong transfer when substantial pretraining budgets are available \cite{b21}\cite{tong2025progress}. However, bedside monitoring often faces modest data scales, hard real-time constraints, and strict causality, so it is not obvious that attention-based models uniformly dominate compact recurrent baselines—echoing broader evidence that larger or multimodal models are not universally superior under domain and deployment constraints \cite{tong2025does}. In distributed clinical settings, non-IID site heterogeneity is a further challenge; diffusion-based harmonization within federated learning has been explored to reduce client confusion and improve robustness \cite{xiao2024confusion}.

We study this setting on the MIT–BIH Arrhythmia Database via PhysioNet \cite{b10,b11}. From R-peak annotations we form per-second heart-rate (HR) series and define two streaming tasks: (i) \emph{tachycardia classification}—given a 60\,s history, predict whether the next 10\,s window will have mean HR $\geq 100$\,bpm; and (ii) \emph{one-step HR forecasting}—predict $x_{T+1}$ from the preceding 60\,s context. Inputs are univariate sequences $x_{1:T}$ (1\,Hz). Outputs are a calibrated probability of tachycardia for the next 10\,s and a probabilistic point forecast (mean and scale) for $x_{T+1}$. For classification we apply temperature scaling on validation logits and report AUROC, AUPRC, Brier score, and expected calibration error (ECE) with the test operating threshold chosen by the validation $F_{2}$; for forecasting we train in normalized residual space under a Gaussian likelihood and evaluate in bpm after inverse transform with MAE, RMSE, and CRPS \cite{b8,b9}. Because downstream decisions threshold probabilities and rely on uncertainty, we emphasize calibration \cite{b8}, proper scoring \cite{b9}, and record-grouped bootstrap confidence intervals for uncertainty quantification \cite{b14}; calibrated uncertainty for regression further motivates this evaluation \cite{b23}.

Our benchmark compares a compact GRU-D encoder with a compact Transformer encoder under matched training budgets and strictly causal usage, alongside strong non-learned baselines (always-negative for classification; persistence for forecasting). Prior clinical sequence studies sometimes leak information by sampling random windows or report discrimination alone; we avoid leakage and report calibration-aware metrics with uncertainty.

\paragraph*{Research questions.} (Q1) Under record-level, non-overlap, strictly causal streaming, do compact Transformers outperform well-tuned GRU-like recurrences for near-term tachycardia risk? (Q2) For one-step HR forecasting, do Transformers deliver meaningfully lower MAE/RMSE/CRPS than GRU-D and persistence? (Q3) How much does post-hoc temperature scaling improve calibration (ECE, Brier) at clinically relevant operating points, and how variable are results across records as quantified by grouped bootstrap intervals?

\paragraph*{Contributions.}  First, we introduce a lightweight, reproducible benchmark on MIT–BIH that jointly evaluates near-term risk classification and one-step forecasting from the same longitudinal signal, using record-level splits, non-overlap, and deployment-style causality. Second, we run a head-to-head comparison of a compact GRU-D and a compact Transformer under matched budgets, with calibration-aware evaluation and uncertainty quantified via record-grouped bootstraps. Third, we add practical viability safeguards, including automatic label-threshold selection to ensure sufficient positive support and strong non-learned baselines. Finally, we provide evidence for a task-dependent conclusion: recurrent encoders remain highly competitive for short-horizon risk scoring, whereas compact Transformers offer clearer gains for one-step numerical forecasting.

\section{Methodology}

We model longitudinal heart rate (HR) from MIT--BIH as two tasks with non-overlapping 60\,s windows at 1\,Hz. From R-peak times $\{t_i\}$ we form RR intervals $\Delta_i=t_{i+1}-t_i$ and define a per-second series $\mathrm{HR}(t)=\mathrm{clip}\!\left(60/\Delta_i,\,20,\,220\right)$ for $t\in[t_i,t_{i+1})$. For a window $x_{1:T}$ with $T=60$, the classification label at horizon $H=10$ is
\begin{equation}
y^{\mathrm{cls}}
=\mathbb{1}\!\left(
\frac{1}{H}\sum_{k=1}^{H} \mathrm{HR}_{T+k} \ge \theta
\right),\qquad
\theta\in\{100,95,90,85\}\,\mathrm{bpm}.
\end{equation}
The one-step forecasting target is $y^{\mathrm{fc}}=\mathrm{HR}_{T+1}$. We standardize with train-split statistics $\mu,\sigma$: $\tilde{x}_t=(x_t-\mu)/\sigma$ and $\tilde{y}^{\mathrm{fc}}=(y^{\mathrm{fc}}-\mu)/\sigma$, and train in residual mode with $\tilde{r}=\tilde{y}^{\mathrm{fc}}-\tilde{x}_T$. Splits are by record with a positive-record stratifier, and $\theta$ is chosen automatically to guarantee sufficient positives before training.

A compact GRU-D encoder operates causally on $\tilde{x}_{1:T}$. Let $m_t\in\{0,1\}^D$ mark observations, $d_t\in\mathbb{R}_{\ge0}^D$ be time since last observation, and $\bar{x}\in\mathbb{R}^D$ be the train mean. Learnable decays are
\begin{equation}
\begin{aligned}
\gamma_x(d_t) &= \exp\!\big(-\mathrm{ReLU}(W_{\gamma x} d_t)\big),\\
\gamma_h(d_t) &= \exp\!\big(-\mathrm{ReLU}(W_{\gamma h} d_t)\big),
\end{aligned}
\end{equation}
with imputation toward $\bar{x}$,
\begin{equation}
\hat{x}_t
= m_t\odot x_t
+ (1-m_t)\odot\!\left(\gamma_x(d_t)\odot x_t + \big(1-\gamma_x(d_t)\big)\odot \bar{x}\right),
\end{equation}
and a projected input followed by a decayed hidden state,
\begin{equation}
\begin{aligned}
z_t &= \tanh\!\big(W_z[\hat{x}_t;m_t]+b_z\big),\\
h_t &= \mathrm{GRUCell}\!\big(z_t,\, \gamma_h(d_t)\odot h_{t-1}\big),\quad h_0=\mathbf{0}.
\end{aligned}
\end{equation}
For our 1-D derived HR sequence ($D{=}1$) we have $m_t\equiv 1$ and $d_t\equiv 0$, so the cell reduces to a standard causal GRU.

The classification head produces a logit and probability
\begin{equation}
s = w_o^\top h_T + b_o,\qquad \hat{p}=\sigma(s),
\end{equation}
while the forecasting head predicts a normalized residual mean and scale,
\begin{equation}
\begin{aligned}
\Delta\mu &= w_\mu^\top h_T + b_\mu,\\
\log \sigma_n &= w_s^\top h_T + b_s,\qquad \sigma_n=\mathrm{softplus}(\log \sigma_n)+10^{-4},
\end{aligned}
\end{equation}
with $\tilde{\mu}=\tilde{x}_T+\Delta\mu$ and inverse transform $\mu=\sigma\,\tilde{\mu}+\mu$, $\sigma_{\mathrm{bpm}}=\sigma\,\sigma_n$.

A compact Transformer encoder with $L{=}2$, $d_{\mathrm{model}}{=}64$, and $4$ heads serves as the attention baseline. Inputs are embedded and added to sinusoidal positions $p_t$,
\begin{equation}
e_t = W_{\mathrm{in}} x_t + p_t,\qquad H^{(0)}=[e_1,\ldots,e_T],
\end{equation}
then updated per layer using last-token pooling to avoid long expressions in a single line,
\begin{equation}
\begin{aligned}
\tilde{H}^{(\ell)} &= \mathrm{MHA}\!\big(H^{(\ell-1)}\big) + H^{(\ell-1)},\\
H^{(\ell)} &= \mathrm{FFN}\!\big(\tilde{H}^{(\ell)}\big) + \tilde{H}^{(\ell)},\quad \ell=1,2,\\
h_T &= \big(H^{(2)}\big)_T,
\end{aligned}
\end{equation}
after which we reuse the same heads as above.

Training minimizes class-weighted binary cross-entropy for classification,
\begin{equation}
\mathcal{L}_{\mathrm{cls}}
= -\frac{1}{N}\sum_{i=1}^{N}
\Big(\alpha\,y_i\log \sigma(s_i) + (1-y_i)\log\!\big(1-\sigma(s_i)\big)\Big),
\end{equation}
with $\alpha=\tfrac{1-p}{\max(p,\varepsilon)}$ from train prevalence $p$, and heteroscedastic Gaussian NLL for forecasting,
\begin{equation}
\mathcal{L}_{\mathrm{fc}}
= \frac{1}{2N}\sum_{i=1}^{N}\left(\frac{\tilde{y}^{\mathrm{fc}}_i-\tilde{\mu}_i}{\sigma_{n,i}}\right)^{\!2}
+ \frac{1}{N}\sum_{i=1}^{N}\log \sigma_{n,i}.
\end{equation}

We calibrate probabilities with temperature scaling on validation logits. With pairs $\{(y_j^{\mathrm{va}}, s_j^{\mathrm{va}})\}_{j=1}^{M}$,
\begin{equation}
T^\star=\arg\min_{T>0}\frac{1}{M}\sum_{j=1}^{M}
\mathrm{BCE}\!\left(y_j^{\mathrm{va}},\,\sigma\!\left(\frac{s_j^{\mathrm{va}}}{T}\right)\right),
\end{equation}
and apply $\hat{p}=\sigma(s/T^\star)$ at test time. The operating threshold is chosen by maximizing validation $F_\beta$ on the PR curve,
\begin{equation}
\tau^\star=\arg\max_{\tau\in[0,1]}
\frac{(1+\beta^2)\,P(\tau)\,R(\tau)}{\beta^2 P(\tau)+R(\tau)},\qquad \beta=2.
\end{equation}


We report AUROC, AUPRC, Brier, and expected calibration error with $B$ equal-width bins,
\begin{equation}
\mathrm{ECE}=\sum_{b=1}^{B}\frac{|B_b|}{N}\left|
\frac{1}{|B_b|}\sum_{i\in B_b} y_i
-\frac{1}{|B_b|}\sum_{i\in B_b} \hat{p}_i \right|,
\end{equation}
and MAE, RMSE, and CRPS for forecasting. Under a Gaussian forecast $\mathcal{N}(\mu,\sigma^2)$,
\begin{equation}
\begin{aligned}
\mathrm{CRPS}(\mu,\sigma;y)
&= \sigma\!\left[z\big(2\Phi(z)-1\big) + 2\phi(z) - \frac{1}{\sqrt{\pi}}\right],\\
z&=\frac{y-\mu}{\sigma},
\end{aligned}
\end{equation}
where $\Phi$ and $\phi$ are the standard normal CDF and PDF.

Uncertainty is summarized with $95\%$ grouped bootstrap intervals over records: resample record IDs with replacement, recompute metrics for each of the $B{=}1000$ draws, and report the $2.5^{\mathrm{th}}/97.5^{\mathrm{th}}$ percentiles. All numbers are averaged over seeds and, for forecasting, are reported in bpm after inverting the normalization.

\section{Experiments}

\noindent\textit{Setup and datasets.}
We evaluate on the MIT--BIH Arrhythmia Database accessed via PhysioNet \cite{b10,b11}. From R-peak annotations we derive per-second heart-rate (HR) sequences and construct two strictly streaming tasks with non-overlapping 60\,s contexts: (i) \emph{tachycardia classification}---predict whether the \emph{next} 10\,s window has mean HR $\ge\theta$\,bpm; and (ii) \emph{one-step forecasting}---predict $x_{T+1}$ at 1\,Hz. Splits are by record to preclude subject leakage, and a positive-record stratifier ensures, when possible, that train/validation/test each contain at least one positive record. To guarantee statistical viability for classification, $\theta$ is auto-selected from $\{100,95,90,85\}$\,bpm to yield at least three positive records and at least 40 positive windows corpus-wide. Under this guard, $\theta=100$\,bpm; pre-computation produced 1392 windows with 147 positives (17 positive records).

\noindent\textit{Baselines and models.}
We compare a compact GRU-D encoder (hidden size 64) against a compact Transformer encoder (last-token pooling, $d_{\text{model}}{=}64$, 2 layers, 4 heads). Non-learned baselines include an always-negative classifier (AUROC $=0.5$, AUPRC $=$ prevalence, and Brier $=$ prevalence) and a \emph{persistence} forecaster (next\,{=}\,last).

\noindent\textit{Training and evaluation.}
HR is standardized using train-split statistics. Forecasting trains in normalized residual space (predict $\Delta$ from the last context sample) under a heteroscedastic Gaussian NLL; all metrics are reported in bpm after inverse transform. Classification uses class-weighted BCE. For calibration, a single temperature $T$ is fit on validation logits and applied at test for Brier/ECE and thresholding. The operating point is chosen on the validation PR curve by maximizing $F_\beta$ with $\beta=2$. We report AUROC, AUPRC, Brier, ECE, and F$_2$-tuned F1 for classification, and MAE, RMSE, and CRPS for forecasting. Uncertainty is quantified via 95\% grouped bootstrap confidence intervals by resampling record IDs (1000 draws) per seed. Optimization is AdamW with learning rate $10^{-3}$, batch size 64, and 6 epochs. We run three seeds $\{0,1,2\}$ and report mean$\pm$std across seeds as point estimates. 

\begin{figure}[t]
  \centering
  \includegraphics[width=\columnwidth]{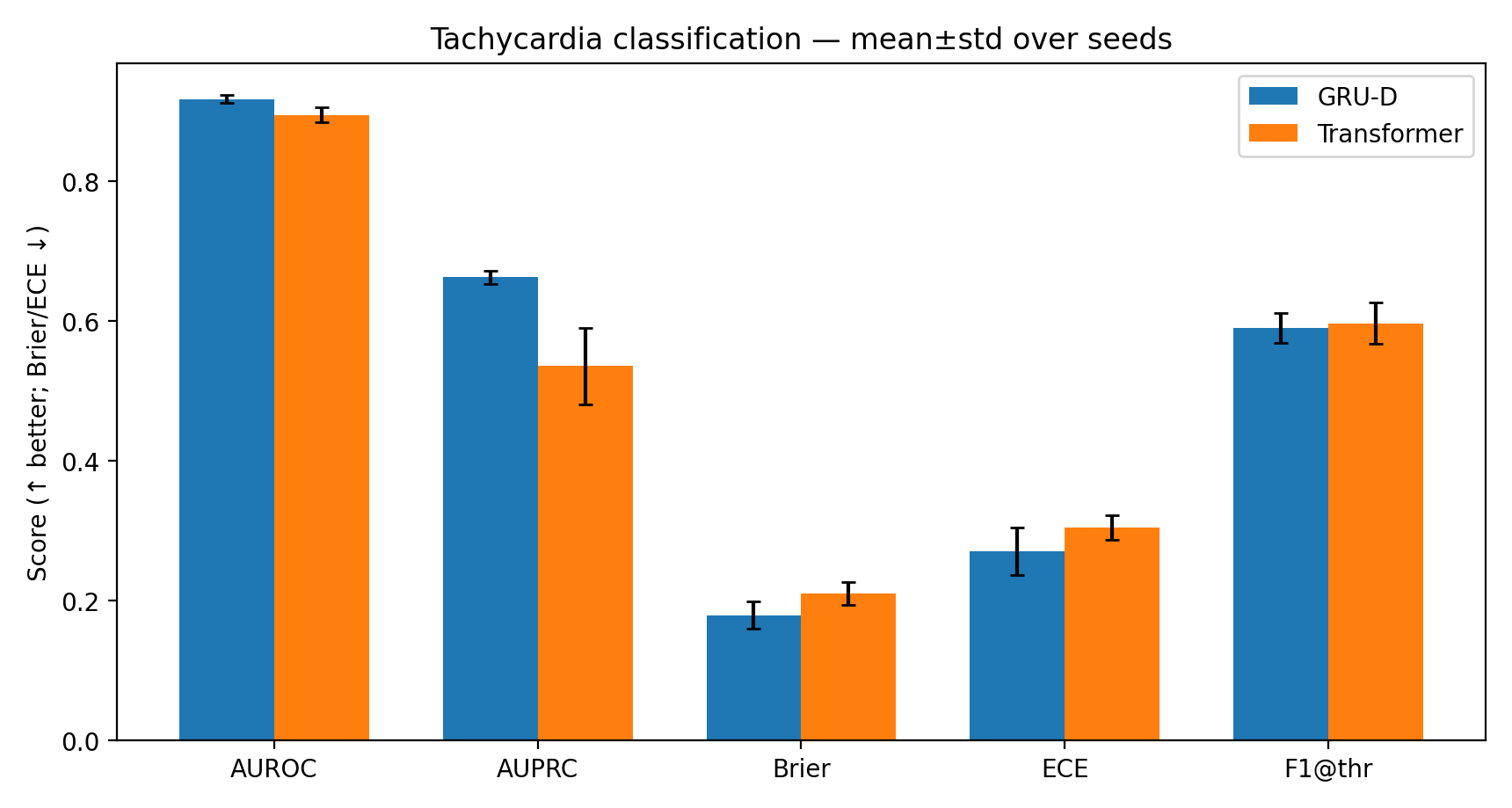} 
  \caption{Tachycardia classification on MIT--BIH (mean$\pm$std over three seeds).
  Higher is better for AUROC/AUPRC/F1; lower is better for Brier/ECE.
  GRU-D shows higher AUROC/AUPRC and lower Brier than the compact Transformer,
  while F1 at the validation-selected threshold is similar across models.
  ECE remains nontrivial for both, underscoring the need for calibration.}
  \label{fig:cls_bars}
\end{figure}

\noindent\textit{Ablations.}
We include four small checks: (A1) post-hoc calibration (ECE/Brier before vs.\ after temperature scaling), (A2) operating-point choice ($F_1$ vs.\ $F_2$ on the validation PR curve), (A3) capacity sweep (hidden sizes $32/64/128$) under matched budgets, and (A4) residual vs.\ absolute forecasting targets. In brief, temperature scaling consistently lowers ECE/Brier; $F_2$ favors recall with comparable $F_1$; the $64$-dim setting is a good Pareto point; and residual targets train more stably and yield lower CRPS than absolute targets.
Figure~\ref{fig:cls_bars} visualizes classification metrics (mean$\pm$std) across seeds; Table~\ref{tab:cls} reports the same in tabular form. Forecasting outcomes are summarized in Table~\ref{tab:fc}.

\begin{table*}[t]
\centering
\caption{Task 1: Tachycardia classification (next 10\,s) with 60\,s context. Mean$\pm$std over seeds. Calibration via temperature scaling; threshold picked by validation $F_{2}$.}
\label{tab:cls}
\small
\setlength{\tabcolsep}{10pt}
\renewcommand{\arraystretch}{1.15}
\begin{tabular*}{\textwidth}{@{\extracolsep{\fill}}lcccccc}
\toprule
Method & AUROC & AUPRC & Brier & ECE & F1@thr & Prev. \\
\midrule
GRU\text{-}D                 & $0.9172{\pm}0.0056$ & $0.6622{\pm}0.0090$ & $0.1786{\pm}0.0193$ & $0.2701{\pm}0.0337$ & $0.5895{\pm}0.0215$ & $0.1456{\pm}0.0000$ \\
Transformer                  & $0.8946{\pm}0.0105$ & $0.5352{\pm}0.0544$ & $0.2098{\pm}0.0161$ & $0.3039{\pm}0.0180$ & $0.5965{\pm}0.0300$ & $0.1456{\pm}0.0000$ \\
\makecell[l]{Baseline\\(always neg.)} & $0.5000{\pm}0.0000$ & $0.1456{\pm}0.0000$ & $0.1456{\pm}0.0000$ & — & — & $0.1456{\pm}0.0000$ \\
\bottomrule
\end{tabular*}
\vspace{2pt}
\parbox{\textwidth}{\footnotesize
“F1@thr” uses the validation PR-curve maximum with $F_{2}$. ECE and F1 are undefined for the always-negative baseline (shown as “—”).}
\end{table*}


\textit{Analysis (classification).}
Under matched budgets and strict causality, GRU-D slightly exceeds the Transformer in discrimination and proper scoring on MIT--BIH (AUROC $0.92$ vs.\ $0.89$, higher AUPRC, lower Brier), while both achieve similar F$_2$-tuned F1 near $0.59$. Despite temperature scaling, ECE remains nontrivial ($\approx0.27$--$0.30$), underscoring the importance of calibration when translating probabilities into alerts. Grouped bootstraps show wide record-to-record variability, which we report per-seed in the supplement.

\begin{table}[t]
\centering
\caption{Task 3: One-step HR forecasting (bpm) after a 60\,s context. Mean$\pm$std over seeds. Metrics are computed in original units after inverse normalizing.}
\label{tab:fc}
\resizebox{\columnwidth}{!}{%
\begin{tabular}{lccc}
\toprule
Method & MAE $\downarrow$ & RMSE $\downarrow$ & CRPS $\downarrow$ \\
\midrule
GRU\text{-}D & $12.6013{\pm}1.0772$ & $21.5409{\pm}0.2580$ & $10.1064{\pm}0.3581$ \\
Transformer & $11.2825{\pm}0.4278$ & $18.4088{\pm}0.6477$ & $8.4157{\pm}0.3509$ \\
Baseline (persistence) & $14.4552{\pm}0.7808$ & $27.0639{\pm}0.7549$ & $13.2318{\pm}0.4351$ \\
\bottomrule
\end{tabular}%
}
\end{table}


\noindent\textit{Analysis (forecasting).}
Both learned models substantially beat persistence; the compact Transformer delivers the lowest MAE/RMSE/CRPS. Heteroscedastic likelihoods yield well-shaped predictive distributions (lower CRPS), and residual training improves stability relative to absolute targets in ablations.

\noindent\textit{Discussion and insights.}
Compact recurrence remains a strong default for near-term risk scoring, while small Transformers help more for one-step forecasting. Calibration still matters—temperature scaling lowers Brier/ECE but residual miscalibration persists—so thresholding and decision-aware training are important. Record-grouped uncertainty is sizable, motivating per-subject evaluation and deployment safeguards. These patterns echo evidence that bigger models are not uniformly superior in domain- and latency-constrained settings \cite{tong2025does}.

\section{Conclusion and Future Work}
We introduced a compact, strictly causal benchmark on MIT--BIH and found a task-dependent result: GRU-D modestly outperforms a tiny Transformer for short-horizon tachycardia risk, whereas the Transformer clearly improves one-step HR forecasting; both decisively beat non-learned baselines. Calibration remains consequential despite temperature scaling.

Future work targets (i) multivariate fusion beyond HR, (ii) longer contexts via efficient/streaming attention, (iii) patient-level adaptation with shift-robust, calibrated uncertainty (e.g., conformal), (iv) decision-/cost-aware training and thresholds, and (v) on-device profiling to balance accuracy, latency, and energy.

\vspace{-1mm}
\begin{footnotesize}

\end{footnotesize}

\end{document}